\documentclass{article}
\usepackage{ijcai18}

\usepackage{times}
\usepackage{xcolor}
\usepackage{soul}
\usepackage[utf8]{inputenc}
\usepackage[small]{caption}
\usepackage{amsmath,epsfig}
\usepackage{times}
\usepackage{epsfig}
\usepackage{graphicx}
\usepackage{amsmath}
\usepackage{dsfont}
\usepackage{amssymb}
\usepackage{url}
\usepackage{amsfonts}
\usepackage{subfigure}
\usepackage{multirow}
\usepackage{makecell}
\usepackage{booktabs}
\usepackage{colortbl}
\usepackage{color}
\title{Rethinking Diversified and Discriminative Proposal Generation\\ for Visual Grounding}

\author{
Zhou Yu$^1$,~
Jun Yu$^1$\thanks{Jun Yu is the corresponding author},~
Chenchao Xiang$^1$,~
Zhou Zhao$^2$,~
Qi Tian$^3$,~
Dacheng Tao$^4$
\\
$^1$ School of Computer Science and Technology, Hangzhou Dianzi University, P. R. China\\
$^2$ College of Computer Science, Zhejiang University, P. R. China\\
$^3$ Department of Computer Science, University of Texas at San Antonio, USA\\
$^4$ UBTECH Sydney AI Centre, SIT, FEIT, University of Sydney, Australia\\
$\{$yuz,~yujun~, ccxiang$\}$@hdu.edu.cn,~
zhaozhou@zju.edu.cn,\\
qi.tian@utsa.edu,~
dacheng.tao@sydney.edu.au
}
\begin{document}

\maketitle

\begin{abstract}
 Visual grounding aims to localize an object in an image referred to by a textual query phrase. Various visual grounding approaches have been proposed, and the problem can be modularized into a general framework: proposal generation, multi-modal feature representation, and proposal ranking. Of these three modules, most existing approaches focus on the latter two, with the importance of proposal generation generally neglected. In this paper, we rethink the problem of what properties make a good proposal generator. We introduce the diversity and discrimination simultaneously when generating proposals, and in doing so propose Diversified and Discriminative Proposal Networks model (DDPN). Based on the proposals generated by DDPN, we propose a high performance baseline model for visual grounding and evaluate it on four benchmark datasets. Experimental results demonstrate that our model delivers significant improvements on all the tested data-sets (e.g., 18.8$\%$ improvement on ReferItGame and 8.2$\%$ improvement on Flickr30k Entities over the existing state-of-the-arts respectively).
\end{abstract}

\section{Introduction}
Recent advances in deep neural networks have helped to solve many challenges in computer vision and natural language processing. These advances have also stimulated high-level research into the connections that exist between vision and language such as visual captioning \cite{donahue2015long,xu2015show}, visual question answering \cite{fukui2016multimodal,yu2017mfb,yu2018beyond} and visual grounding \cite{rohrbach2016grounding,chen2017query}.

\begin{figure}
\begin{center}
\includegraphics[width=0.49\textwidth]{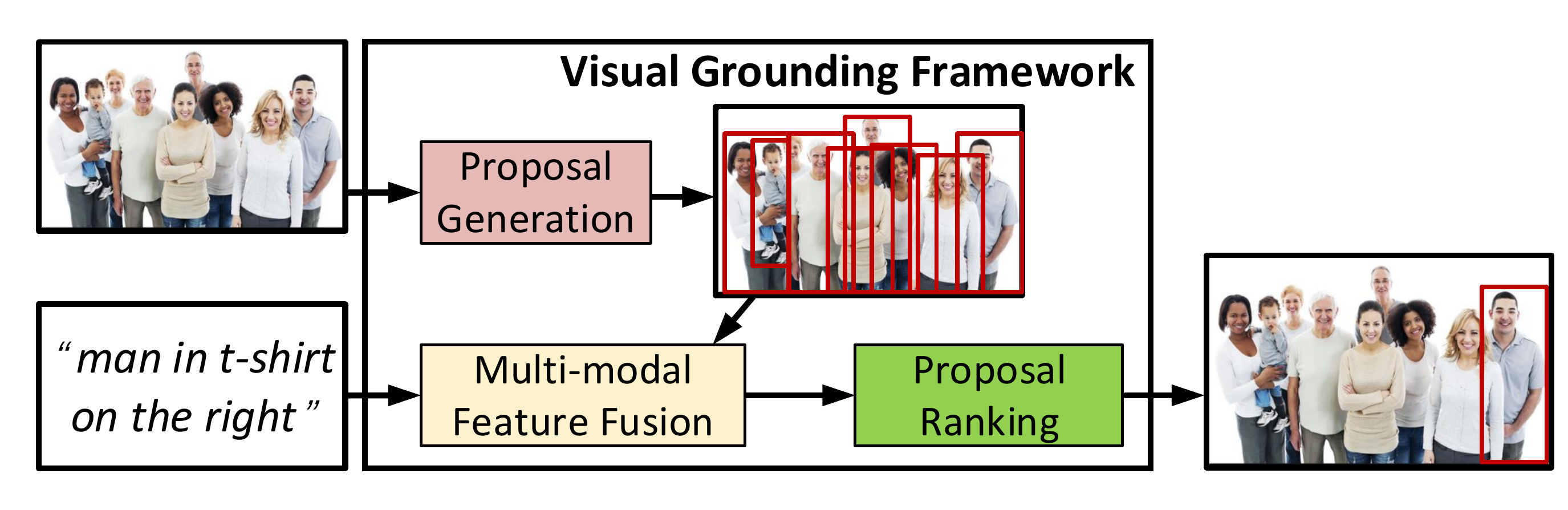}
\vspace{-15pt}
\caption{The general framework of the visual grounding task. Given an arbitrary image and an open-ended query phrase as the inputs, the visual grounding model outputs the predicted bounding box of the referred object. }
\label{fig:vg_framework}
\end{center}
\vspace{-20pt}
\end{figure}

Visual grounding (a.k.a., referring expressions) aims to localize an object in an image referred to by a textual query phrase. It is a challenge task that requires a fine-grained understanding of the semantics of the image and the query phrase and the ability to predict the location of the object in the image. The visual grounding task is a natural extension of object detection \cite{ren2015faster}. While object detection aims to localize all possible pre-defined objects, visual grounding introduces a textual query phrase to localize only one best matching object from an open vocabulary.

Most existing visual grounding approaches \cite{rohrbach2016grounding,fukui2016multimodal,hu2016natural} can be modularized into the general framework shown in Figure \ref{fig:vg_framework}. Given an input image, a fixed number of object proposals are first generated using a proposal generation model and visual features for each proposal are then extracted. Meanwhile, the input query is encoded by a language model (e.g., LSTM \cite{hochreiter1997long}) that outputs a textual feature vector. The textual and visual features are then integrated by a multi-modal fusion model and fed into a proposal ranking model to output the location of the proposal with the highest ranking score.

Of the three modules shown in Figure \ref{fig:vg_framework}, most existing visual grounding approaches focus on designing robust multi-modal feature fusion models. SCRC \cite{hu2016natural} and phrase-based CCA \cite{plummer2015flickr30k} learn the multi-modal common space via canonical correlation analysis (CCA) and a recurrent neural network (RNN), respectively. MCB learns a compact bilinear pooling model to fuse the multi-modal features, to obtain a more discriminative fused feature \cite{fukui2016multimodal}. Further, some approaches have explored designing different loss functions to help improve the accuracy of the final object localization. Rohrbach \emph{et al.} use an auxiliary reconstruction loss function to regularize the model training \cite{rohrbach2016grounding}. Wu \emph{et al.} propose a model that gradually refine the predicted bounding box via reinforcement learning \cite{wu2017end}.

Compared with the aforementioned two modules, proposal generation has been less thoroughly investigated. Most visual grounding approaches usually use class-agnostic models (e.g., selective search \cite{uijlings2013selective} or data-driven region proposal networks (RPNs) \cite{ren2015faster} trained on specific datasets) to generate object proposals and extract visual features for each proposal \cite{fukui2016multimodal,hu2016natural,rohrbach2016grounding,chen2017query,li2017deep}. Although many visual grounding approaches have been proposed, \emph{what makes a good proposal generator for visual grounding} remains uncertain.

Choosing the optimal number of generated proposals for visual grounding is difficult. If the number is small, recall of the true objects is limited, which seriously influencing visual grounding performance. If the number is large, recall is satisfactory but it does not necessarily improve visual grounding performance, as it increases the difficulty for accurate prediction of the proposal ranking model. Here, we rethink this problem and suggest that the ideal generated proposal set should contain as many different objects as possible but be of a relatively small size. To achieve this goal, we introduce \emph{diversity} and \emph{discrimination} simultaneously when generating proposals, and in doing so propose Diversified and Discriminative Proposal Networks (DDPN).

Based on the proposals generated by DDPN, we propose a high performance baseline model to verify the effectiveness of the DDPN for visual grounding. Our model uses simple feature concatenation as the multi-modal fusion model, and trained with two novel loss functions: Kullback-Leibler Divergence (KLD) loss with soft labels to penalize the proposals while capturing contextual information of the generated proposals, and smoothed $L_1$ regression loss to refine the proposal bounding boxes.

The main contributions of this study are as follows: 1) we analyze the limitations of existing proposal generators for visual grounding and propose DDPN to generate high-quality proposals; 2) based on DDPN, we propose a high performance baseline model trained with two novel losses; and 3) by way of extensive ablation studies, we evaluate our models on four benchmark datasets. Our experimental results demonstrate that our model delivers a significant improvement on all the tested datasets.

\section{The Diversified and Discriminative Proposal Networks (DDPN)}\label{sec:ddrp}

\begin{figure}
\begin{center}
\subfigure[Diversified Proposals] {\includegraphics[width=0.48\linewidth]{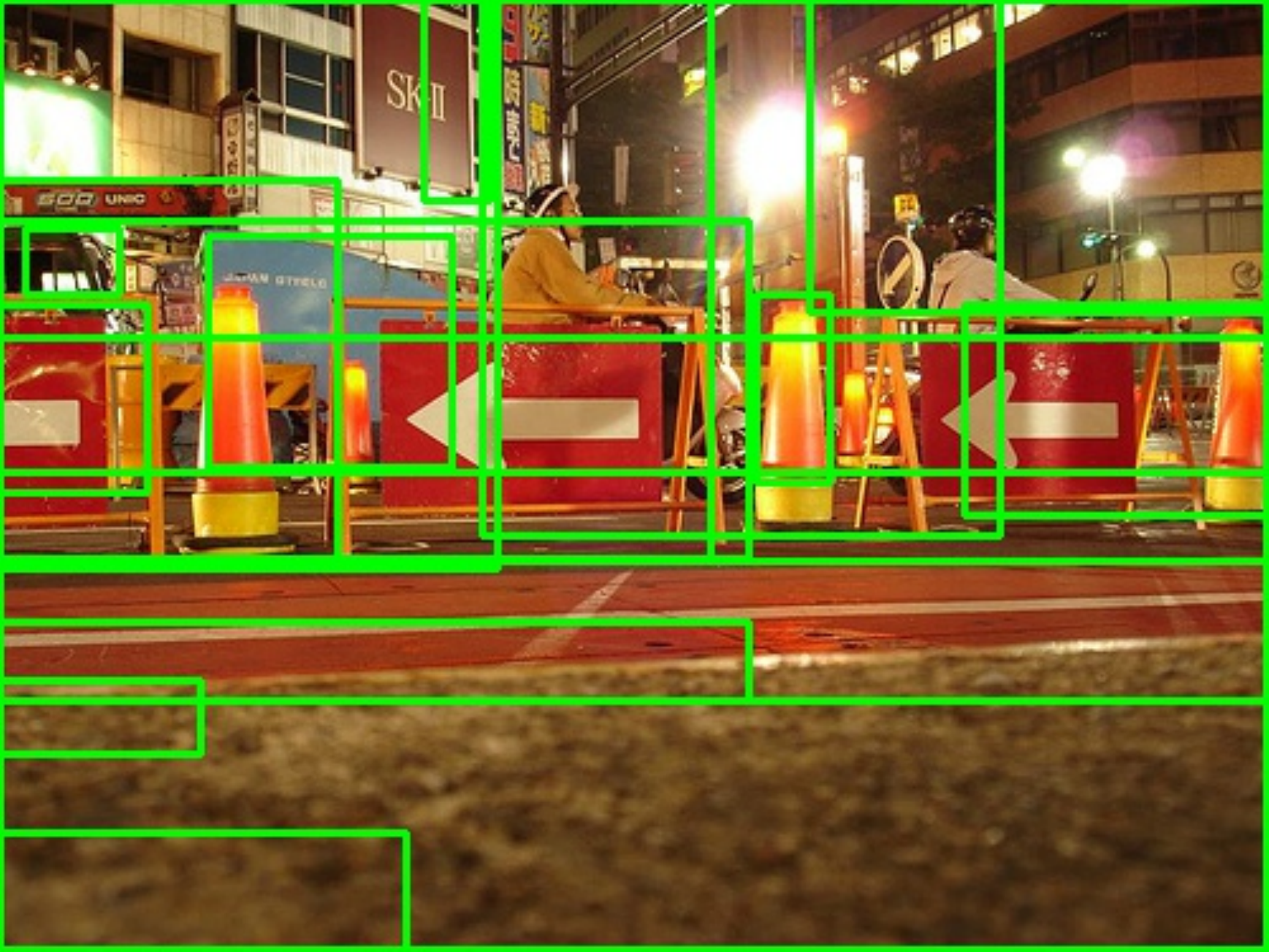}\label{fig:div_example}}
\subfigure[Discriminative Proposals] {\includegraphics[width=0.48\linewidth]{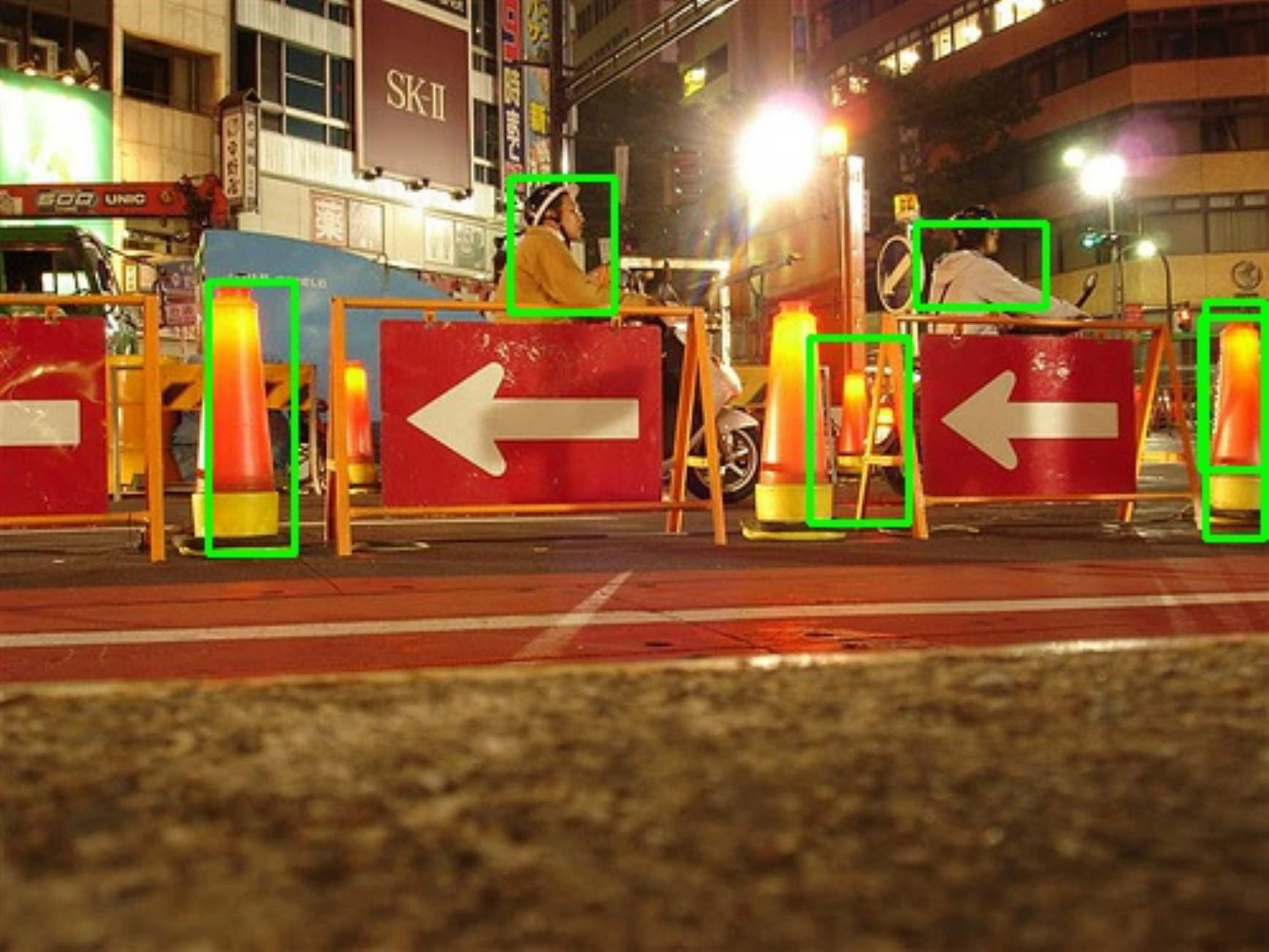}\label{fig:discri_example}}
\vspace{-10pt}
\caption{Examples of diversified proposals and discriminative proposals.}
\label{fig:ddpg_examples}
\end{center}
\vspace{-20pt}
\end{figure}

For visual grounding, we propose that the ideal generated proposals should be diversified and discriminative simultaneously: 1) the proposals of all images should be diversified to detect objects from open-vocabulary classes (see Figure \ref{fig:div_example}). As the objects to be localized during visual grounding may vary significantly, proposals only covering objects from a small set of classes will be a bottleneck for the model that follows and limit final visual grounding performance; 2) the proposals of an individual image should be discriminative to guarantee that the proposals and visual features accurately represent the true semantic (see Figure \ref{fig:discri_example}).

In practice, learning a proposal generator that fully meets these two properties is challenging. The diversified property requires a \emph{class-agnostic} object detector, while the discriminative property requires a \emph{class-aware} detector. Since class-aware object detectors are usually trained on datasets with a small set of object classes (e.g., 20 classes for PASCAL VOC \cite{everingham2010pascal} or 80 classes for COCO \cite{lin2014microsoft}), directly using the output bounding boxes of the detection model as the proposals for visual grounding may compromise diversity. As a trade-off, most existing visual grounding approaches only consider diversity and ignore the discrimination during proposal generation. Specifically, they use the {class-agnostic} object detector (e.g., Selective Search \cite{uijlings2013selective}) to generate proposals and then extract the visual features for the proposals using a pre-trained model \cite{rohrbach2016grounding,hu2016natural,lin2014microsoft,fukui2016multimodal} or a fine-tuned model \cite{chen2017query}. However, when there is only a limited number of proposals, they may have the following limitations: 1) they may not be accurate enough to localize the corresponding objects; 2) they may fail to localize small objects; and 3) they may contain some noisy information (i.e., meaningless proposals).

To overcome the drawbacks inherent in the existing approaches, we relax the diversity constraint to a relatively large set of object classes (e.g., more than 1k classes) and propose a simple solution by training a class-aware object detector that can detect a large number of object classes as an approximation. We choose the commonly used class-aware model Faster R-CNN \cite{ren2015faster} as the object detector, and train the model on the Visual Genome dataset which contains a large number of object classes and extra attribute labels \cite{krishna2016visual}. We call our scheme Diversified and Discriminative Proposal Networks (DDPN).

Following \cite{anderson2017up-down}, we clean and filter of the training data in Visual Genome by truncating the low frequency classes and attributes. Our final training set contains 1,600 object classes and 400 attribute classes. To train this model, we initialize Faster R-CNN with the CNN model (e.g., VGG-16 \cite{simonyan2014very} or ResNet-101 \cite{he2015deep}) pre-trained on ImageNet. We then train Faster R-CNN on the processed Visual Genome datasets. Slightly different from the standard Faster R-CNN model with four losses, we add an additional loss to the last fully-connected layer to predict attribute classes in addition to object classes. Specifically, we concatenate the fully-connected feature with a learned embedding of the ground-truth object class and feed this into an additional output layer defining a softmax distribution over the attribute classes.

Note that although DDPN is similar to \cite{anderson2017up-down}, we have a different motivation. In their approach, Faster R-CNN is used to build bottom-up attention to provide high-level image understanding. In this paper, we use Faster R-CNN to learn diversified and discriminative proposals for visual grounding.

\section{Baseline Model for Visual Grounding}
\begin{figure}
\begin{center}
\includegraphics[width=0.49\textwidth]{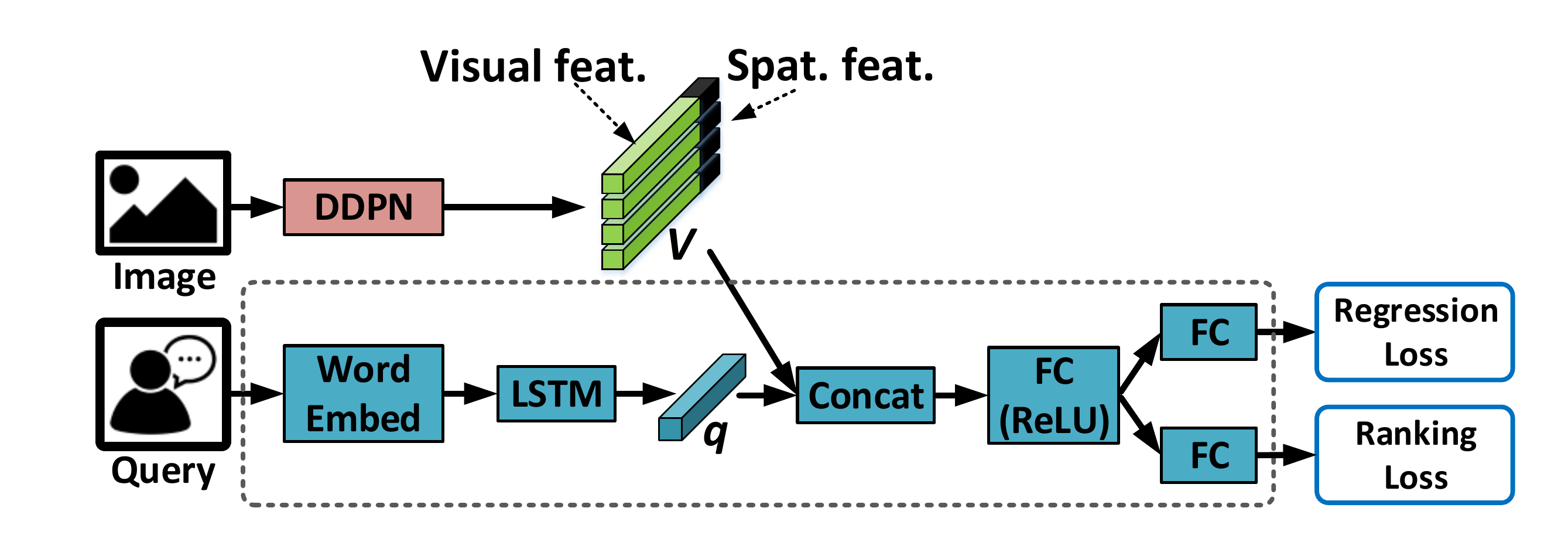}
\vspace{-15pt}
\caption{The flowchart of our visual grounding model during the training stage. DDPN corresponds to the model described in section \ref{sec:ddrp}. The parameters in our model (within the dashed box) is optimized via back-propagation with the regression and ranking losses.}
\label{fig:baseline}
\end{center}
\vspace{-15pt}
\end{figure}

We first introduce a high performance baseline model for visual grounding based on DDPN. Given an image-query pair, our model is trained to predict the bounding boxes of the referred object in the image. The flowchart of our model is shown in Figure \ref{fig:baseline}.

\begin{equation}\label{eq:lstm}
q = \mathrm{LSTM}(\mathrm{Embed}(p))
\end{equation}

For an input image, we first use the DDPN model to extract the top-$N$ proposals with diversity and discrimination. Each proposal is represented as a visual feature (the last fully-connected layer after RoI pooling) $v_{vis}\in\mathbb{R}^{d_v}$ and a 5-D spatial feature $v_{spat}=[x_{tl}/W, y_{tl}/H, x_{br}/W, y_{br}/H, wh/WH]$ proposed in \cite{yu2016modeling}. We concatenate $v_{vis}$ (with $L_2$ normalization) and $v_{spat}$ to obtain the visual feature $v\in\mathbb{R}^{d_v+5}$ for each region proposal. For $N$ region proposals, the visual features are defined as $V=[v_1,...,v_N]$.

For an input textual query $p=[w_1,w_2,...,w_t]$ describing the object in the image, $t$ is the query length, and each word $w_i$ is represented as a one-hot feature vector referring to the index of the word in a large vocabulary. As the length varies for different queries, we use the LSTM network to encode variable-length queries \cite{hochreiter1997long}. Specifically, we first learn a word embedding layer to embed each word into a $d_e$-dimensional latent space. The sequence of the embedded word features is then fed to a one-layer LSTM network with $d_q$ hidden units. We extract the feature of the last word from the LSTM network as the output feature $q\in\mathbb{R}^{d_q}$ for the query phrase $p$. The encoding procedure for the query phrases is represented as follows.

Once we have obtained the query feature $q\in\mathbb{R}^{d_q}$ and visual features $V=[v_1,...,v_N]\in\mathbb{R}^{d_v\times N}$, we can fuse the multi-modal features with a feature fusion model and output the integrated features $F=[f_1,...,f_N]$.

To better demonstrate the capacity of the visual features, we do not introduce a complex feature fusion model (e.g., bilinear pooling) and only use simple feature concatenation followed by a fully-connected layer for fusion. For each pair $\{q, v\}$, the output feature $f$ is obtained as follows.
\begin{equation}\label{eq:feature_fusion}
f=\varphi(W_f^T(q||v)+b_f)
\end{equation}
where $W_f\in\mathbb{R}^{(d_v+d_q)\times d_o}$ and $b_f\in\mathbb{R}^{d_o}$ are the weights and bias respectively, $||$ denotes concatenation of vectors, $f\in\mathbb{R}^{d_o}$ is the fused feature. $\varphi(\cdot)$ is the ReLU activation function.

\subsection{Learning Objective}
Given the fused feature $f\in F$, we use an additional fully-connected layer to predict the score $s\in\mathbb{R}$ for feature $f$.
\begin{equation}\label{eq:score_pred}
s= W_s^Tf+b_s
\end{equation}
where $W_s\in\mathbb{R}^{o}$ and $b_s\in\mathbb{R}$ are the weights and bias, respectively. For $N$ proposals, we can obtain their scores $S=[s_1,...,s_N]\in\mathbb{R}^N$ using Eq.(\ref{eq:score_pred}). We apply softmax normalization to $S$ to make it satisfy a score distribution.

The problem now becomes how to define the loss function to make the predicted ranking scores $S$ consistent with their ground-truth ranking scores $S^*$. In most existing visual grounding approaches \cite{fukui2016multimodal,hu2016natural,chen2017query}, the ground-truth ranking scores $S^*=[s_1^*,...,s_N^*]\in\{0,1\}^N$ are defined as a one-hot vector in which the only element is set to 1 when the corresponding proposal most overlaps with the ground-truth bounding box (i.e., the largest IoU score) and 0 otherwise. Based on the one-hot single label, softmax cross-entropy loss is used to learning the ranking model.

Slightly different to their strategy, we use the \emph{soft} label distribution as the ground-truth. We calculate the IoU scores of all proposals w.r.t. the ground-truth bounding box and assign the IoU score of the $i$-th proposal to $s_i^*$ if the IoU score is larger than a threshold $\eta$ and 0 otherwise. In this way, we obtain the real-value label vector $S^*=[s_1^*,...,s^*_N]\in\mathbb{R}^N$. To make $S^*$ satisfy a probability distribution, $S^*$ is $L_1$-normalized to guarantee $\sum_i^Ns^*_i=1$. Accordingly, we use Kullback-Leibler Divergence (KLD) as our ranking function for model optimization and to make the predicted score distribution $S$ and ground-truth label distribution $S^*$ as close as possible.
\begin{equation}\label{eq:kld_loss}
L_{\textrm{rank}}= \frac{1}{N}\sum\limits_i^Ns_i^*\mathrm{log}(\frac{s_i^*}{s_i})
\end{equation}

The benefits of using {soft} labels are three-fold: 1) except for the max-overlapping proposal, other proposals may also contain useful information, that provides contextual knowledge of the ground-truth; 2) the soft label can be seen as a model regularization strategy, as introduced in \cite{szegedy2016rethinking} as label smoothing; and 3) during testing, the predicted bounding box is considered to be correct when its IoU score with the ground-truth is larger than a threshold $\eta$. Therefore, optimizing the model with soft labels applies consistency to training and testing and may improve visual grounding performance.

Although using the KLD loss in Eq.(\ref{eq:kld_loss}) can capture the proposals' contextual information, the accuracies of the region proposals themselves could be a performance bottleneck for visual grounding. In the case that all the generated region proposals do not overlap with the ground-truth bounding box, the grounding accuracy will be zero. Inspired by the strategy used in Faster R-CNN \cite{ren2015faster}, we append an additional fully-connected layer on top of fused feature $f$ and add a regression layer to refine the proposal coordinates.
\begin{equation}\label{eq:reg_pred}
t= W_t^Tf+b_t
\end{equation}
where $t\in\mathbb{R}^4$ corresponds to the coordinates of the refined bounding box, and $W_t\in\mathbb{R}^{d_o\times 4}$ and $b_t\in\mathbb{R}^4$ are the weights and bias respectively. Accordingly, the smoothed $L_1$ regression loss is defined as follows to penalize the difference between $t$ and its ground-truth bounding box coordinates $t^*$ for all the proposals.
\begin{equation}\label{eq:reg_loss}
L_{\textrm{reg}}= \frac{1}{N}\sum\limits_i^N \mathrm{smooth}_{L_1}(t_i^*, t_i)
\end{equation}
where $\mathrm{smooth}_{L_1}(x,y)$ is the smoothed $L_1$ loss function for input features $x$ and $y$ \cite{ren2015faster}.
It is worth noting that although the proposals generated by DDPN has been refined by a regression layer in the Faster R-CNN model, the regression layer here has a different meaning: it acts as a domain adaptation function to align the semantics of the source domain (i.e., the Visual Genome dataset) and the target domain (i.e., the specific visual grounding dataset).

The overall loss for our model is defined as:
\begin{equation}\label{eq:localization_loss}
L= L_{\textrm{rank}}+\gamma L_{\textrm{reg}}
\end{equation}
where $\gamma$ is a hyper-parameter to balance the two terms.

%\subsection{Implementation Details}

\section{Experiments}
We evaluate our approach on four benchmark datasets: Flickr30K Entities \cite{plummer2015flickr30k}, ReferItGame \cite{kazemzadeh2014referitgame}, RefCOCO and RefCOCO+ \cite{yu2016modeling}. These are all commonly used benchmark datasets for visual grounding.

\subsection{Implementation Details}
We use the same hyper-parameters as in \cite{anderson2017up-down} to train the DDPN model. We train DDPN with different CNN backbones (VGG-16 and ResNet-101). Both models are trained for up to 30 epochs, which takes two GPUs 2$\sim$3 weeks to finish.

Based on the optimized DDPN model, we train our visual grounding model with the hyper-parameters listed as follows. The loss weight $\gamma$ is set to 1 for all experiments. The dimensionality of the visual feature $d_v$ is 2048 (for ResNet-101) or 4096 (for VGG-16), the dimensionality of the word embedding feature $d_e$ is 300, the dimensionality of the output feature of the LSTM network $d_q$ is 1024, the dimensionality of the fused feature $d_o$ is 512, the threshold of IoU scores $\eta$ is 0.5, and the number of proposals $N$ is 100. During training, all the weight parameters, including the word embedding layer, the LSTM network and the multi-modal fusion model are initialized using the {xavier} method. We use the Adam solver to train the model with $\beta_1=0.9$, $\beta_2=0.99$. The base learning rate is set to 0.001 with an exponential decay rate of 0.1. The mini-batch size is set to 64. All the models are trained up to 10k iterations. During testing, we feed forward the network and output the scores for all the proposals before picking the proposal with the highest score and using its refined bounding box as the final output.

%All experiments are implemented with the \emph{Caffe} toolbox \cite{jia2014caffe} and performed on a workstation with four NVIDIA TitanXP GPUs.

\subsection{Datasets}
\subsubsection{Flickr30K Entities}
Flickr30K Entities contains 32k images, and 275k bounding boxes and 360k query phrases. Some phrases in the dataset may correspond to multiple boxes. In such cases, we merge the boxes and use their union as the ground-truth \cite{rohrbach2016grounding}. We use the standard split in our setting, i.e., 1k images for validation, 1k for testing, and 30k for training.

\subsubsection{ReferItGame}
ReferItGame contains over 99k bounding boxes and 130k query phrases from 20k images. Each bounding box is associated with 1-3 query phrases. We use the same data split as in \cite{rohrbach2016grounding}, namely 10k images for testing, 9k for training and 1k for validation.

\subsubsection{RefCOCO and RefCOCO+}
Both the RefCOCO and RefCOCO+ datasets utilize images from the COCO dataset \cite{lin2014microsoft}. Both of them consist of 142k query phrases and 50k bounding boxes from 20k images. The difference between the two datasets is that the query phrases in RefCOCO+ does not contain any location word, which is more difficult to understand the query intent. The datasets are split into four sets: train, validation, testA and testB. The images in testA contain multiple people, while images in testB contain objects in other categories.
%We use the original split provided in \cite{yu2016modeling}.
\begin{figure*}
\begin{center}
\subfigure[SS] {\includegraphics[width=0.24\linewidth]{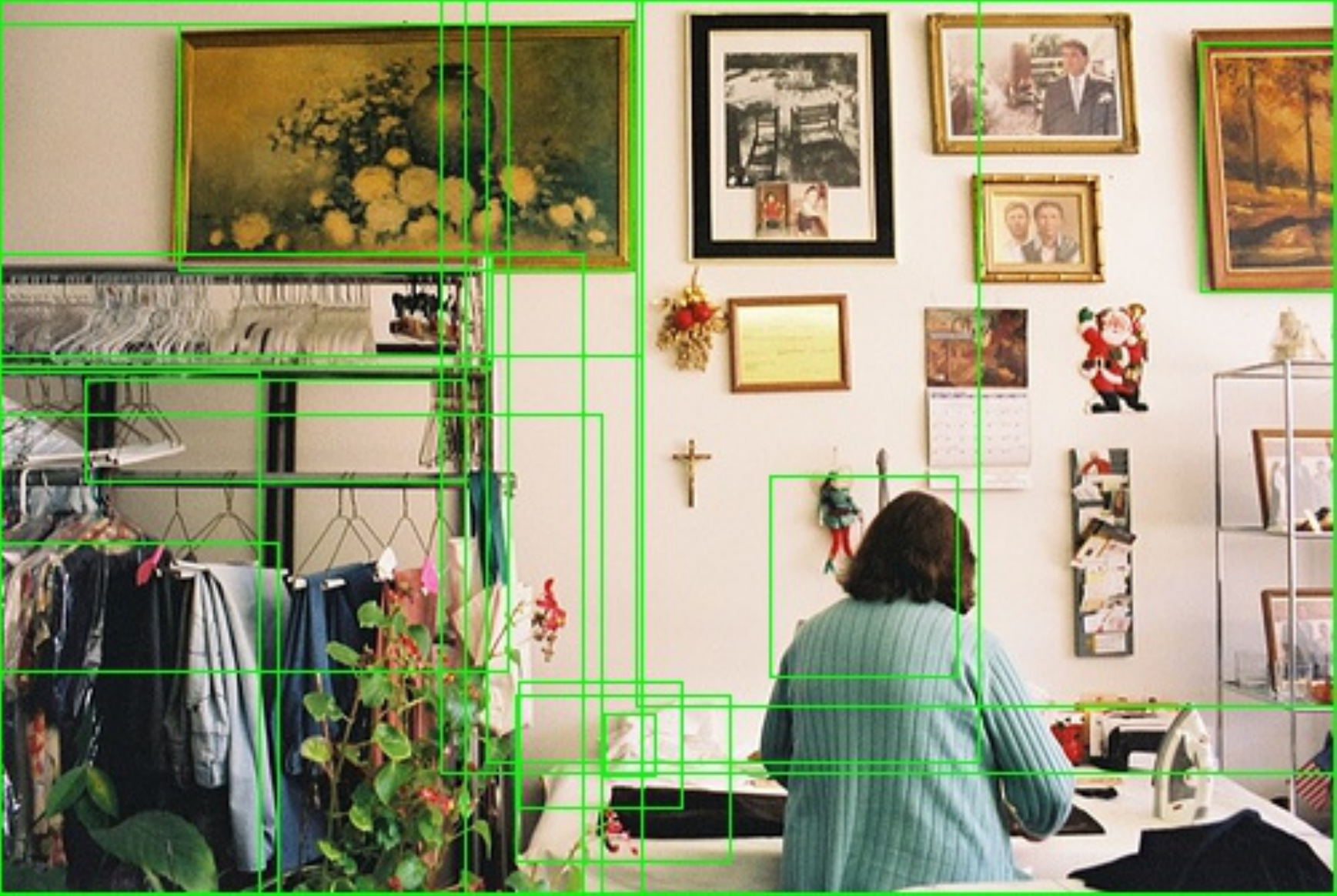}\label{fig:div_ss}}
\subfigure[RPN (w/o ft)] {\includegraphics[width=0.24\linewidth]{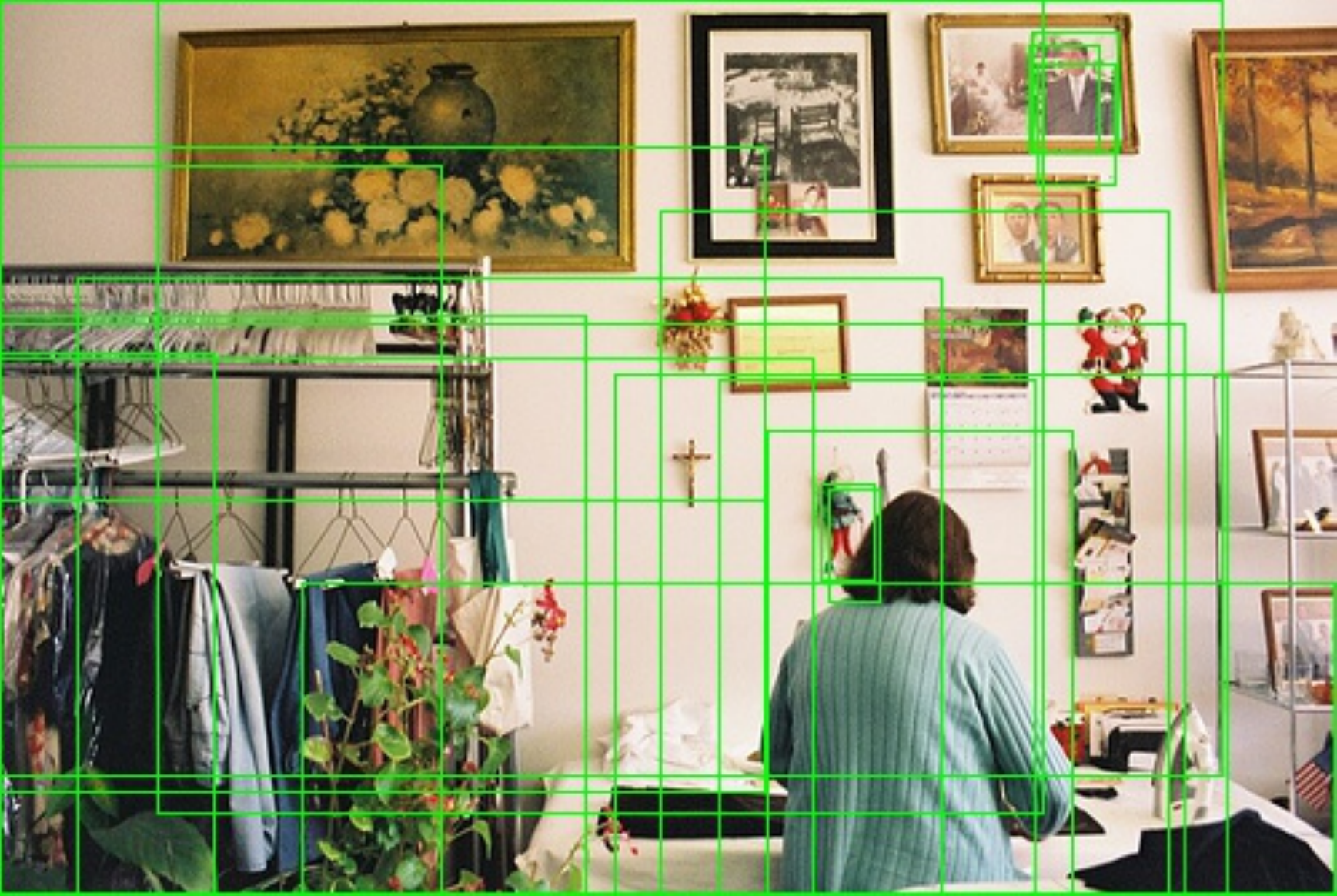}\label{fig:div_rpn_noft}}
\subfigure[RPN (w/ ft)] {\includegraphics[width=0.24\linewidth]{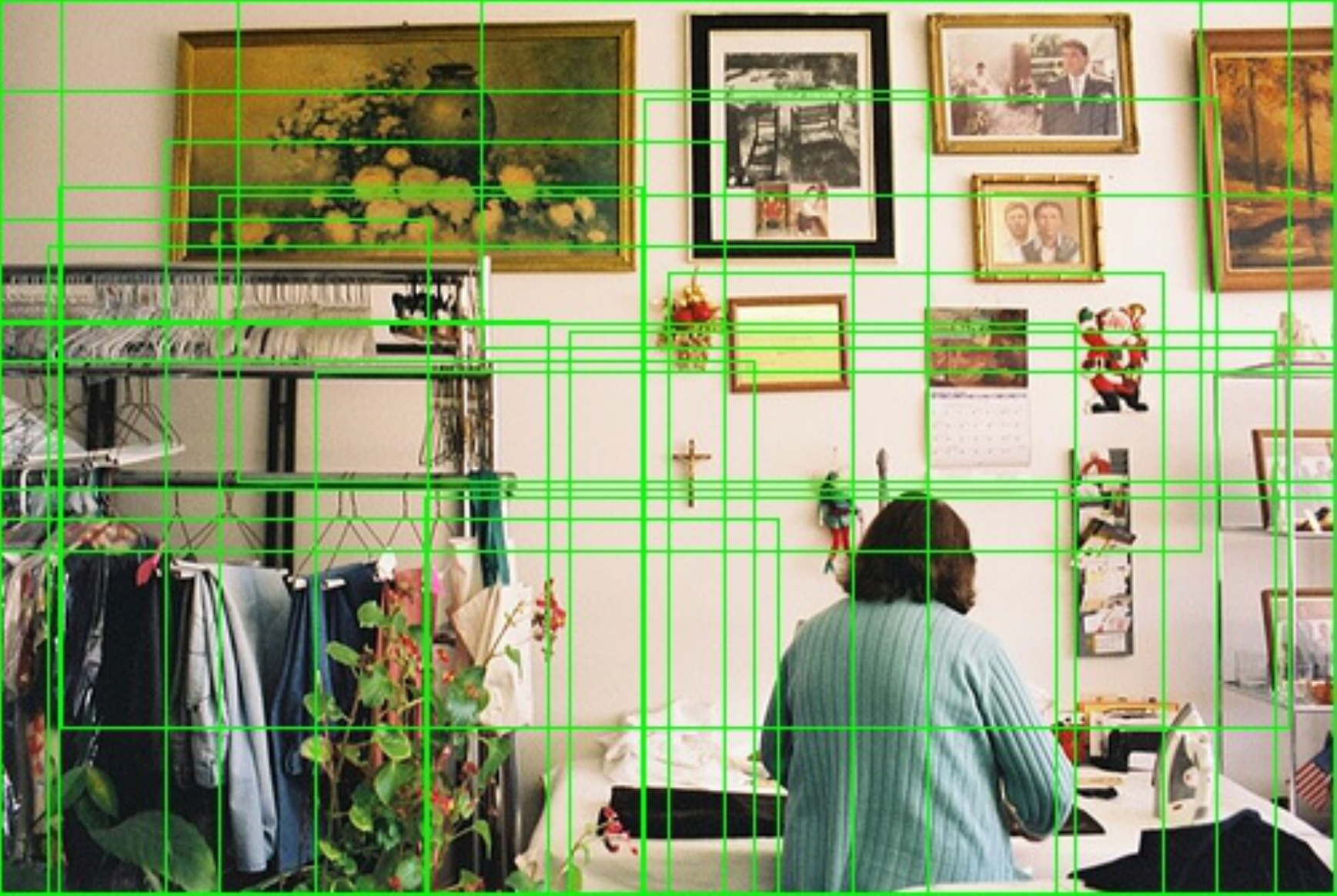}\label{fig:div_rpn_ft}}
\subfigure[DDPN] {\includegraphics[width=0.24\linewidth]{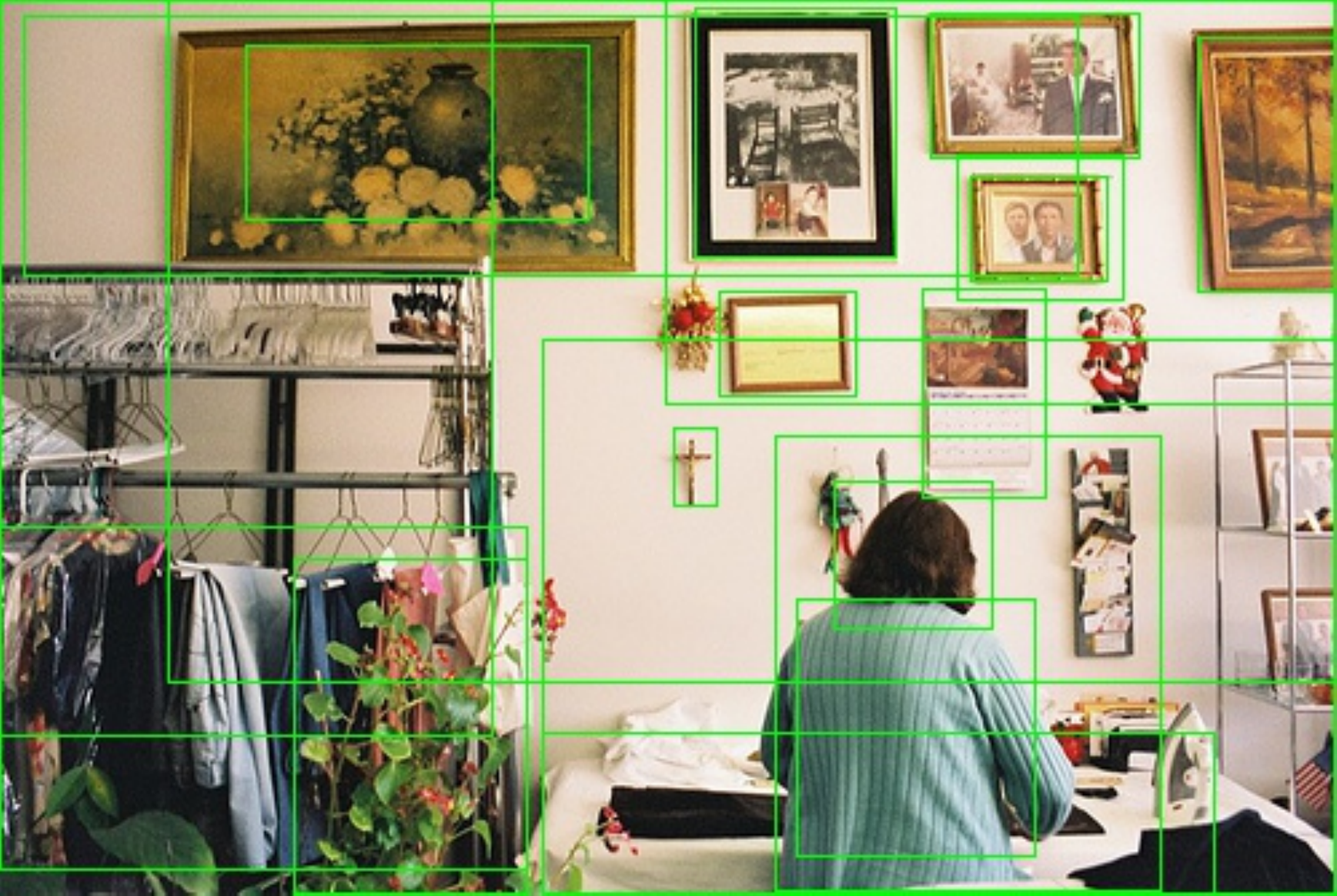}\label{fig:div_ddpg}}
\vspace{-10pt}
\caption{The generated region proposals with different proposal generators on Flickr30k: (a) selective search (SS); (b) region proposal network (RPN) trained on PASCAL VOC 07; (c) the RPN model in (b) with fine-tuning on Flickr30k Entities; (d) our diversified and discriminative proposal networks (DDPN). For better visualization, we only show the top-20 proposals for each model w.r.t. their proposal scores in descending order. }
\label{fig:div_comparasion}
\end{center}
\vspace{-15pt}
\end{figure*}
\subsection{Experimental Setup}
\subsubsection{Evaluation Metric}
For all datasets, we adopt accuracy as the evaluation metric, which is defined as the percentage in which the predicted bounding box overlaps with the ground-truth of IoU $> 0.5$.
\subsubsection{Compared Methods}
We compare our approach with state-of-the-art visual grounding methods: SCRC \cite{hu2016natural}, GroundeR \cite{rohrbach2016grounding}, MCB \cite{fukui2016multimodal}, QRC \cite{chen2017query}, and the approaches of Li \emph{et al.} \cite{li2017deep}, Wu \emph{et al.} \cite{wu2017end} and Yu \emph{et al.} \cite{yu2017joint}.
%For GroundeR, we make comparisons with its supervised learning scenario since achieves the best performance.

\subsection{Ablation Studies}
We also perform the following ablation experiments on Flickr30k Entities to verify the efficacy of DDPN, and the loss functions to train our baseline visual grounding model.

\subsubsection{Effects of Diversity and Discrimination in Proposal Generation}

We define the {discrimination} score ($S_{\mathrm{DIS.}}$) and the {diversity} score ($S_{\mathrm{DIV.}}$) in Eq.(\ref{eq:ddscore}) to evaluate the proposal generators.

\begin{equation}
\begin{array}{rcl}\label{eq:ddscore}
S_{\mathrm{DIS.}} &=& \frac{1}{M}\sum\limits_{i}^M \mathop{\mathrm{max}}\limits_j~\sigma(t_i^j,t_i^*)\\
S_{\mathrm{DIV.}} &=& 1/\frac{1}{M}\sum\limits_{i}^M\sum\limits_j^N~\sigma(t_i^j,t_i^*)\\
\end{array}
\end{equation}
where $M$ is the number of query-image samples in the test set, $N$ is the number of proposals per sample, and $t_i^j$ indicates the $j$-th proposal in the $i$-th sample. For the $i$-th sample, the indicator function $\sigma(t_i^j,t_i^*)=1$ if a proposal $t_i^j $ covers its corresponding ground-truth bounding box $t_i^*$ with IoU $>$ 0.5 and 0 otherwise. $S_{\mathrm{DIS.}}$ is the discrimination score which is defined as the percentage of samples with at least one proposal covering its ground-truth objects. A larger discrimination score leads to more accurate proposals. $S_{\mathrm{DIV.}}$ is the diversity score defined as the \emph{reciprocal} of the average number of covered proposals per ground-truth object. A larger diversity score leads to fewer redundant proposals and a more balanced proposal distribution. A good proposal generator should have high $S_{\mathrm{DIV.}}$ and $S_{\mathrm{DIS.}}$ simultaneously.

In Table \ref{table:aba_ddpg}, we report the $S_{\mathrm{DIV.}}$ and $S_{\mathrm{DIS.}}$, as well as the visual grounding accuracies of different models by replacing DDPN (shown in Figure \ref{fig:baseline}) with other proposal generators. The results show that: 1) SS and RPN (w/o ft) models have similar discriminative scores, while the diversity score of RPN (w/o ft) is smaller than that of SS. This reflects the fact that SS is more diversified compared to the data-driven RPN model pre-trained on PASCAL VOC 07 \cite{everingham2010pascal}; 2) after fine-tuning on Flickr30k Entities, the RPN (w/ ft) model obtains a better discriminative score than RPN (w/o ft). However, the diversity score is reduced, indicating that RPN (w/ ft) fits Flickr30k Entities at the expense of model generalizability; 3) DDPN significantly outperforms other approaches for both discriminative and diversity scores. To further demonstrate the effect of DDPN, we illustrate proposals for an image using the aforementioned four models in Figure \ref{fig:div_comparasion}, and the visualized results show that DDPN generates proposals of higher quality than other approaches; 4) DDPN significantly improve the visual grounding performance compared to other proposal generation models, which proves our hypothesis that both diversity and discrimination are of great significance for visual grounding performance.

\subsubsection{Effects of Different Losses on Visual Grounding}
In Table \ref{table:aba_loss_fusion}, we report the performance of our model variants trained with different losses. It can be seen that: 1) training with the KLD loss leads to a 1.2$\sim$1.7-point improvement over the models with classical single-label softmax loss. This verifies our hypothesis that using soft labels captures contextual information and enhances the model's capacity. Compared to the context modeling strategy proposed by \cite{chen2017query}, which exploits the ground-truth bounding boxes within an image from multiple queries, our strategy is more general and easier to implement; 2) training with an additional regression loss to refine the proposals leads to a 4.7$\sim$5.2-point improvement.
We believe this will become a commonly-used strategy in future visual grounding studies.

\begin{table}
\centering
\caption{The discrimination scores ($S_{\mathrm{DIS.}}$), the diversity scores ($S_{\mathrm{DIV.}}$) and the visual grounding accuracies of different proposal generators evaluated on the test set of Flickr30k Entities. All the models use the same VGG-16 backbone model (except SS which does not introduce the learned model). Both the RPN (w/o ft) and RPN (w/ ft) models are pre-trained on PASCAL VOC 07 while RPN (w/ ft) is additionally fine-tuned on Flickr30k Entities.}
\label{table:aba_ddpg}
\vspace{-5pt}
\small
\begin{tabular}{c|cccc}
\toprule
Metrics &SS & RPN (w/o ft) & RPN (w/ ft) &  DDPN\\
\midrule
 $S_{\mathrm{DIS.}}$ & 0.78 & 0.77 & 0.85  & \textbf{0.92} \\
 $S_{\mathrm{DIV.}}$ & 0.16 & 0.14 & 0.12 &  \textbf{0.20}\\
\midrule
 Accuracy ($\%$) &57.9&55.5&59.5&\textbf{70.0}\\
\bottomrule
\end{tabular}
\vspace{-5pt}
\end{table}

\begin{table}
\centering
\caption{The visual grounding accuracies of the variants with different loss functions on Flickr30k Entities. All the variants use the same DDPN model with VGG-16 backbone.}
\vspace{-5pt}
\small
\label{table:aba_loss_fusion}
\begin{tabular}{ccc|cc}
\toprule
\multicolumn{3}{c|}{{\makecell{Different losses}}} & \multirow{2}{*}{\makecell{Accuracy ($\%$)}} & \multirow{2}{*}{Improv. ($\%$)}\\
$L_{\textrm{softmax}}$&$L_{\textrm{kld}}$&$L_{\textrm{reg}}$& \\
\hline
$\checkmark$&&& 63.6 & -\\
%\rowcolor{mygray}
&$\checkmark$&& 64.8 & 1.2\\
$\checkmark$&&$\checkmark$& 68.3 & 4.7\\
%\rowcolor{mygray}
&$\checkmark$&$\checkmark$& \textbf{70.0} & 6.4\\
\bottomrule
\end{tabular}
\vspace{-10pt}
\end{table}

\begin{table*}
\centering
\caption{The accuracies of different methods on RefCOCO and RefCOCO+ datasets}\label{table:refcoco}
\vspace{-5pt}
\small
\begin{tabular}{c|ccc|ccc}
\toprule
 & \multicolumn{3}{c|}{RefCOCO} & \multicolumn{3}{c}{RefCOCO+}\\
Methods & TestA ($\%$) & TestB ($\%$) & Validation ($\%$) & TestA ($\%$) & TestB ($\%$) & Validation ($\%$) \\
\midrule
SCRC \cite{hu2016natural} & 18.5 & 20.2 & 19.0 & 14.4 & 13.3 & 13.7 \\
Wu \emph{et al.} \cite{wu2017end} & 54.8 & 41.6 & 48.2 & 40.4 & 22.8 & 31.9\\
Li \emph{et al.} \cite{li2017deep} & 55.2 &  46.2 & 52.4 & 45.2 & 32.2 & 40.2 \\
Yu \emph{et al.} \cite{yu2017joint} & 73.7 &  65.0 & 69.5 & 60.7 & 48.8 & 55.7 \\
\midrule
Ours (VGG-16) & 76.9& 67.5 & 73.4& 67.0 & 50.2 & 60.1 \\
Ours (ResNet-101) & \textbf{80.1} & \textbf{72.4} & \textbf{76.8} &  \textbf{70.5} & \textbf{54.1} & \textbf{64.8} \\
\bottomrule
\end{tabular}
\vspace{-10pt}
\end{table*}

\begin{figure*}
\begin{center}
\includegraphics[width=0.96\linewidth]{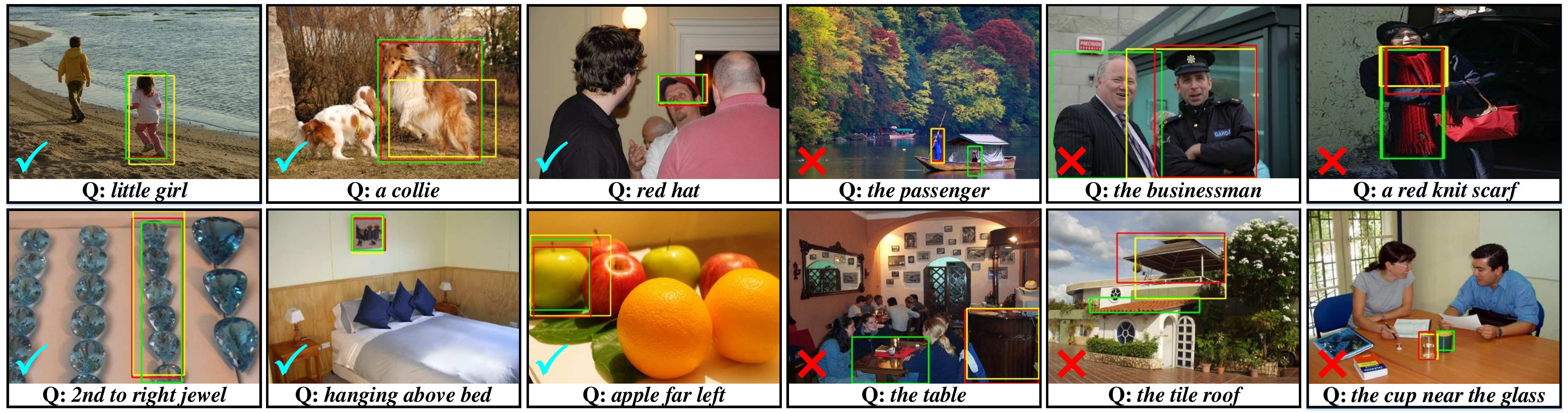}
\vspace{-5pt}
\caption{The examples in Flickr30k Entities (the 1st row) and ReferItGame (the 2nd row) datasets. The ground-truth (green), the top-ranked predicted proposal (yellow) and the refined final prediction (red) are visualized respectively. The last three columns shows the examples with incorrect predictions (IoU $\leq$ 0.5). Best viewed in color.}\label{fig:visualization}
\end{center}
\vspace{-20pt}
\end{figure*}

\subsection{Comparisons with State-of-the-Art}

We next compare our models to current state-of-the-art approaches. We report the results of our approach with DDPN of two backbones, namely VGG-16 \cite{simonyan2014very} and ResNet-101 \cite{he2015deep}.

Tables \ref{table:refcoco}, \ref{table:flickr30k} and \ref{table:referit} show the comparative results on RefCOCO, RefCOCO+, Flickr30k Entities and ReferItGame, respectively. We note the following: 1) with the same CNN backbone (i.e., VGG-16), our model achieves an absolute improvement of 3.2 points on RefCOCO (testA), 6.3 points on RefCOCO+ (testA), 4.9 points on Flickr30k Entities and 16.1 points on ReferItGame, respectively. The improvement is primarily due to the high-quality proposals generated by DDPN, and the loss functions we use for our visual grounding model; 2) by replacing VGG-16 with ResNet-101 as the backbone for DDPN, all results improve by 3$\sim$4 points, illustrating that the representation capacity of DDPN significantly influences the visual grounding performance.
\begin{table}
\centering
\caption{The accuracies of different methods on Flickr30k Entities}\label{table:flickr30k}
\vspace{-5pt}
\small
\begin{tabular}{c|c}
\toprule
 Methods & \makecell{Accuracy ($\%$)} \\
 \midrule
SCRC \cite{hu2016natural}& 27.8\\
GroundeR \cite{rohrbach2016grounding}& 47.8 \\
MCB \cite{fukui2016multimodal}& 48.7 \\
QRC \cite{chen2017query}& 65.1 \\
\midrule
Ours (VGG-16) & 70.0\\
Ours (ResNet-101)& \textbf{73.3} \\
\bottomrule
 \end{tabular}
 \vspace{-10pt}
\end{table}

\subsection{Qualitative Results}
We visualize some visual grounding results of our model with ResNet-101 backbone model on Flickr30k Entities (the first row) and ReferItGame (the second row) in Figure \ref{fig:visualization}. It can be seen that our approach achieves good visual grounding performance, and is able to handle the small and fine-grained objects. Moreover, the bounding box regression helps refine the results, rectifying some inaccurate proposals. Finally, our approach still has some limitations, especially when faced with complex queries or confused visual objects. These observations are useful to guide further improvements for visual grounding in the future.
\begin{table}
\centering
\caption{The accuracies of different methods on ReferItGame}\label{table:referit}
\vspace{-5pt}
\small
\begin{tabular}{c|c}
\toprule
 Methods & \makecell{Accuracy ($\%$)}\\
 \midrule
 SCRC \cite{hu2016natural} & 17.9\\
 GroundeR \cite{rohrbach2016grounding}& 28.5 \\
 MCB \cite{fukui2016multimodal}& 28.9 \\
 Wu \emph{et al.} \cite{wu2017end} & 36.8 \\
 QRC \cite{chen2017query}& 44.1 \\
 Li \emph{et al.} \cite{li2017deep} & 44.2\\
\midrule
Ours (VGG-16) & 60.3\\
Ours (ResNet-101) & \textbf{63.0} \\
\bottomrule
\end{tabular}
\vspace{-10pt}
\end{table}
\section{Conclusions and Future Work}
In this paper, we interrogate the proposal generation for visual grounding and in doing so propose Diversified and Discriminative Proposal Networks (DDPN) to produce high-quality proposals. Based on the proposals and visual features extracted from the DDPN model, we design a high performance baseline for visual grounding trained with two novel losses: KLD loss to capture the contextual information of the proposals and regression loss to refine the proposals. We conduct extensive experiments on four benchmark datasets and achieve significantly better results on all datasets.

Since the models studied here represent the baseline, there remains significant room for improvement, for example by introducing a more advanced backbone model for DDPN or introducing a more powerful multi-modal feature fusion model such as bilinear pooling.
%\appendix
\section*{Acknowledgments}
This work was supported in part by National Natural Science Foundation
of China under Grant 61702143, Grant 61622205, Grant 61472110 and Grant 61602405, and in part by the Zhejiang Provincial Natural Science Foundation of China under Grant LR15F020002, in part by the Australian Research Council Projects under Grant FL-170100117 and Grant DP-180103424.
%\section{\LaTeX{} and Word Style Files}\label{stylefiles}

%\newpage
\bibliographystyle{named}
\bibliography{ijcai18.bbl}

\begin{thebibliography}{}

\bibitem[\protect\citeauthoryear{Anderson \bgroup \em et al.\egroup
  }{2017}]{anderson2017up-down}
Peter Anderson, Xiaodong He, Chris Buehler, Damien Teney, Mark Johnson, Stephen
  Gould, and Lei Zhang.
\newblock Bottom-up and top-down attention for image captioning and visual
  question answering.
\newblock {\em arXiv preprint arXiv:1707.07998}, 2017.

\bibitem[\protect\citeauthoryear{Chen \bgroup \em et al.\egroup
  }{2017}]{chen2017query}
Kan Chen, Rama Kovvuri, and Ram Nevatia.
\newblock Query-guided regression network with context policy for phrase
  grounding.
\newblock {\em ICCV}, 2017.

\bibitem[\protect\citeauthoryear{Donahue \bgroup \em et al.\egroup
  }{2015}]{donahue2015long}
Jeffrey Donahue, Lisa Anne~Hendricks, Sergio Guadarrama, Marcus Rohrbach,
  Subhashini Venugopalan, Kate Saenko, and Trevor Darrell.
\newblock Long-term recurrent convolutional networks for visual recognition and
  description.
\newblock In {\em CVPR}, pages 2625--2634, 2015.

\bibitem[\protect\citeauthoryear{Everingham \bgroup \em et al.\egroup
  }{2010}]{everingham2010pascal}
Mark Everingham, Luc Van~Gool, Christopher~KI Williams, John Winn, and Andrew
  Zisserman.
\newblock The pascal visual object classes (voc) challenge.
\newblock {\em IJCV}, 88(2):303--338, 2010.

\bibitem[\protect\citeauthoryear{Fukui \bgroup \em et al.\egroup
  }{2016}]{fukui2016multimodal}
Akira Fukui, Dong~Huk Park, Daylen Yang, Anna Rohrbach, Trevor Darrell, and
  Marcus Rohrbach.
\newblock Multimodal compact bilinear pooling for visual question answering and
  visual grounding.
\newblock {\em arXiv preprint arXiv:1606.01847}, 2016.

\bibitem[\protect\citeauthoryear{He \bgroup \em et al.\egroup
  }{2016}]{he2015deep}
Kaiming He, Xiangyu Zhang, Shaoqing Ren, and Jian Sun.
\newblock Deep residual learning for image recognition.
\newblock {\em CVPR}, pages 770--778, 2016.

\bibitem[\protect\citeauthoryear{Hochreiter and
  Schmidhuber}{1997}]{hochreiter1997long}
Sepp Hochreiter and J{\"u}rgen Schmidhuber.
\newblock Long short-term memory.
\newblock {\em Neural computation}, 9(8):1735--1780, 1997.

\bibitem[\protect\citeauthoryear{Hu \bgroup \em et al.\egroup
  }{2016}]{hu2016natural}
Ronghang Hu, Huazhe Xu, Marcus Rohrbach, Jiashi Feng, Kate Saenko, and Trevor
  Darrell.
\newblock Natural language object retrieval.
\newblock In {\em CVPR}, pages 4555--4564, 2016.

\bibitem[\protect\citeauthoryear{Kazemzadeh \bgroup \em et al.\egroup
  }{2014}]{kazemzadeh2014referitgame}
Sahar Kazemzadeh, Vicente Ordonez, Mark Matten, and Tamara~L Berg.
\newblock Referitgame: Referring to objects in photographs of natural scenes.
\newblock In {\em EMNLP}, pages 787--798, 2014.

\bibitem[\protect\citeauthoryear{Krishna \bgroup \em et al.\egroup
  }{2017}]{krishna2016visual}
Ranjay Krishna, Yuke Zhu, Oliver Groth, Justin Johnson, Kenji Hata, Joshua
  Kravitz, Stephanie Chen, Yannis Kalantidis, Li-Jia Li, David~A Shamma, et~al.
\newblock Visual genome: Connecting language and vision using crowdsourced
  dense image annotations.
\newblock {\em IJCV}, 123(1):32--73, 2017.

\bibitem[\protect\citeauthoryear{Li \bgroup \em et al.\egroup
  }{2017}]{li2017deep}
Jianan Li, Yunchao Wei, Xiaodan Liang, Fang Zhao, Jianshu Li, Tingfa Xu, and
  Jiashi Feng.
\newblock Deep attribute-preserving metric learning for natural language object
  retrieval.
\newblock In {\em ACM Multimedia}, pages 181--189, 2017.

\bibitem[\protect\citeauthoryear{Lin \bgroup \em et al.\egroup
  }{2014}]{lin2014microsoft}
Tsung-Yi Lin, Michael Maire, Serge Belongie, James Hays, Pietro Perona, Deva
  Ramanan, Piotr Doll{\'a}r, and C~Lawrence Zitnick.
\newblock Microsoft coco: Common objects in context.
\newblock In {\em ECCV}, pages 740--755, 2014.

\bibitem[\protect\citeauthoryear{Plummer \bgroup \em et al.\egroup
  }{2015}]{plummer2015flickr30k}
Bryan~A Plummer, Liwei Wang, Chris~M Cervantes, Juan~C Caicedo, Julia
  Hockenmaier, and Svetlana Lazebnik.
\newblock Flickr30k entities: Collecting region-to-phrase correspondences for
  richer image-to-sentence models.
\newblock In {\em ICCV}, pages 2641--2649, 2015.

\bibitem[\protect\citeauthoryear{Ren \bgroup \em et al.\egroup
  }{2015}]{ren2015faster}
Shaoqing Ren, Kaiming He, Ross Girshick, and Jian Sun.
\newblock Faster r-cnn: Towards real-time object detection with region proposal
  networks.
\newblock In {\em NIPS}, pages 91--99, 2015.

\bibitem[\protect\citeauthoryear{Rohrbach \bgroup \em et al.\egroup
  }{2016}]{rohrbach2016grounding}
Anna Rohrbach, Marcus Rohrbach, Ronghang Hu, Trevor Darrell, and Bernt Schiele.
\newblock Grounding of textual phrases in images by reconstruction.
\newblock In {\em ECCV}, pages 817--834, 2016.

\bibitem[\protect\citeauthoryear{Simonyan and
  Zisserman}{2014}]{simonyan2014very}
Karen Simonyan and Andrew Zisserman.
\newblock Very deep convolutional networks for large-scale image recognition.
\newblock {\em arXiv preprint arXiv:1409.1556}, 2014.

\bibitem[\protect\citeauthoryear{Szegedy \bgroup \em et al.\egroup
  }{2016}]{szegedy2016rethinking}
Christian Szegedy, Vincent Vanhoucke, Sergey Ioffe, Jon Shlens, and Zbigniew
  Wojna.
\newblock Rethinking the inception architecture for computer vision.
\newblock In {\em CVPR}, pages 2818--2826, 2016.

\bibitem[\protect\citeauthoryear{Uijlings \bgroup \em et al.\egroup
  }{2013}]{uijlings2013selective}
Jasper~RR Uijlings, Koen~EA Van De~Sande, Theo Gevers, and Arnold~WM Smeulders.
\newblock Selective search for object recognition.
\newblock {\em IJCV}, 104(2):154--171, 2013.

\bibitem[\protect\citeauthoryear{Wu \bgroup \em et al.\egroup
  }{2017}]{wu2017end}
Fan Wu, Zhongwen Xu, and Yi~Yang.
\newblock An end-to-end approach to natural language object retrieval via
  context-aware deep reinforcement learning.
\newblock {\em arXiv preprint arXiv:1703.07579}, pages 3518--3024, 2017.

\bibitem[\protect\citeauthoryear{Xu \bgroup \em et al.\egroup
  }{2015}]{xu2015show}
Kelvin Xu, Jimmy Ba, Ryan Kiros, Kyunghyun Cho, Aaron~C Courville, Ruslan
  Salakhutdinov, Richard~S Zemel, and Yoshua Bengio.
\newblock Show, attend and tell: Neural image caption generation with visual
  attention.
\newblock In {\em ICML}, volume~14, pages 77--81, 2015.

\bibitem[\protect\citeauthoryear{Yu \bgroup \em et al.\egroup
  }{2016}]{yu2016modeling}
Licheng Yu, Patrick Poirson, Shan Yang, Alexander~C Berg, and Tamara~L Berg.
\newblock Modeling context in referring expressions.
\newblock In {\em ECCV}, pages 69--85, 2016.

\bibitem[\protect\citeauthoryear{Yu \bgroup \em et al.\egroup
  }{2017a}]{yu2017joint}
Licheng Yu, Hao Tan, Mohit Bansal, and Tamara~L Berg.
\newblock A joint speaker-listener-reinforcer model for referring expressions.
\newblock In {\em CVPR}, pages 7282--7290, 2017.

\bibitem[\protect\citeauthoryear{Yu \bgroup \em et al.\egroup
  }{2017b}]{yu2017mfb}
Zhou Yu, Jun Yu, Jianping Fan, and Dacheng Tao.
\newblock Multi-modal factorized bilinear pooling with co-attention learning
  for visual question answering.
\newblock {\em ICCV}, pages 1821--1830, 2017.

\bibitem[\protect\citeauthoryear{Yu \bgroup \em et al.\egroup
  }{2018}]{yu2018beyond}
Zhou Yu, Jun Yu, Chenchao Xiang, Jianping Fan, and Dacheng Tao.
\newblock Beyond bilinear: Generalized multi-modal factorized high-order
  pooling for visual question answering.
\newblock {\em IEEE Transactions on Neural Networks and Learning Systems},
  2018.
\newblock doi:10.1109/TNNLS.2018.2817340.

\end{thebibliography}

\end{document}